\documentclass[runningheads]{llncs}

\usepackage{eccv}

\usepackage{eccvabbrv}

\usepackage{graphicx}
\usepackage{booktabs}
\usepackage{multirow}
\usepackage{booktabs}
\usepackage{makecell}
\usepackage[table]{xcolor}
\usepackage{pifont}
\usepackage{caption}
\newcommand{\greencheck}{\textcolor{green!60!black}{\ding{51}}}
\newcommand{\redcross}{\textcolor{red!80!black}{\ding{55}}}
\usepackage[accsupp]{axessibility}  

\usepackage{hyperref}

\usepackage{orcidlink}

\begin{document}

\title{\textsc{SIMS}plat: Language-Aligned 4D Gaussian Splatting for Driving Scenario Generation} 

\titlerunning{SIMSplat}

\author{
Sung-Yeon Park\inst{1} \and
Adam Lee\inst{2} \and
Juanwu Lu\inst{1} \and
Can Cui\inst{1} \and
Luyang Jiang\inst{1} \and
Rohit Gupta\inst{3} \and
Kyungtae Han\inst{3} \and
Ahmadreza Moradipari\inst{3} \and
Ziran Wang\inst{1}
}

\authorrunning{S.-Y. Park et al.}

\institute{
Purdue University
\and
University of California, Berkeley
\and
Toyota InfoTech Labs \\
\email{\{sungyeon,ziran\}@purdue.edu}
}

\maketitle



\label{sec:abstract}


\begin{abstract}
Driving scene manipulation using real-world sensor data has emerged as a promising alternative to traditional driving simulators. Despite advances in language control and neural scene representations, existing methods treat grounding, editing, and simulation as loosely connected stages, relying on heuristic object localization, manual guidance, and single-agent validation—thereby constraining semantic expressiveness and hindering scalable, reactive scenario generation. We introduce SIMSplat, a driving scene editor built on scene-graph-based 4D Gaussian Splatting augmented with language-aligned features. By embedding appearance, motion, and location semantics directly into Gaussian scene-graph nodes, SIMSplat makes reconstructed scenes queryable through free-form natural language, bridging language understanding to object-level editing and multi-agent simulation within a single framework. Building on this language-grounded scene graph, SIMSplat supports diverse edits including fine-grained pedestrian manipulation, while a multi-agent path refinement module propagates changes across all agents to ensure reactive, physically plausible simulations. The pipeline further integrates with Vision-Language Models for automated scenario mining. Experiments show that SIMSplat more than doubles baseline grounding accuracy, achieves the highest task completion rate, and produces the lowest failure rates across diverse driving scenarios.
\end{abstract}

\section{Introduction}
\label{sec:intro}

\begin{figure*}[!t]
    \centering
    \includegraphics[width=1.0\textwidth]{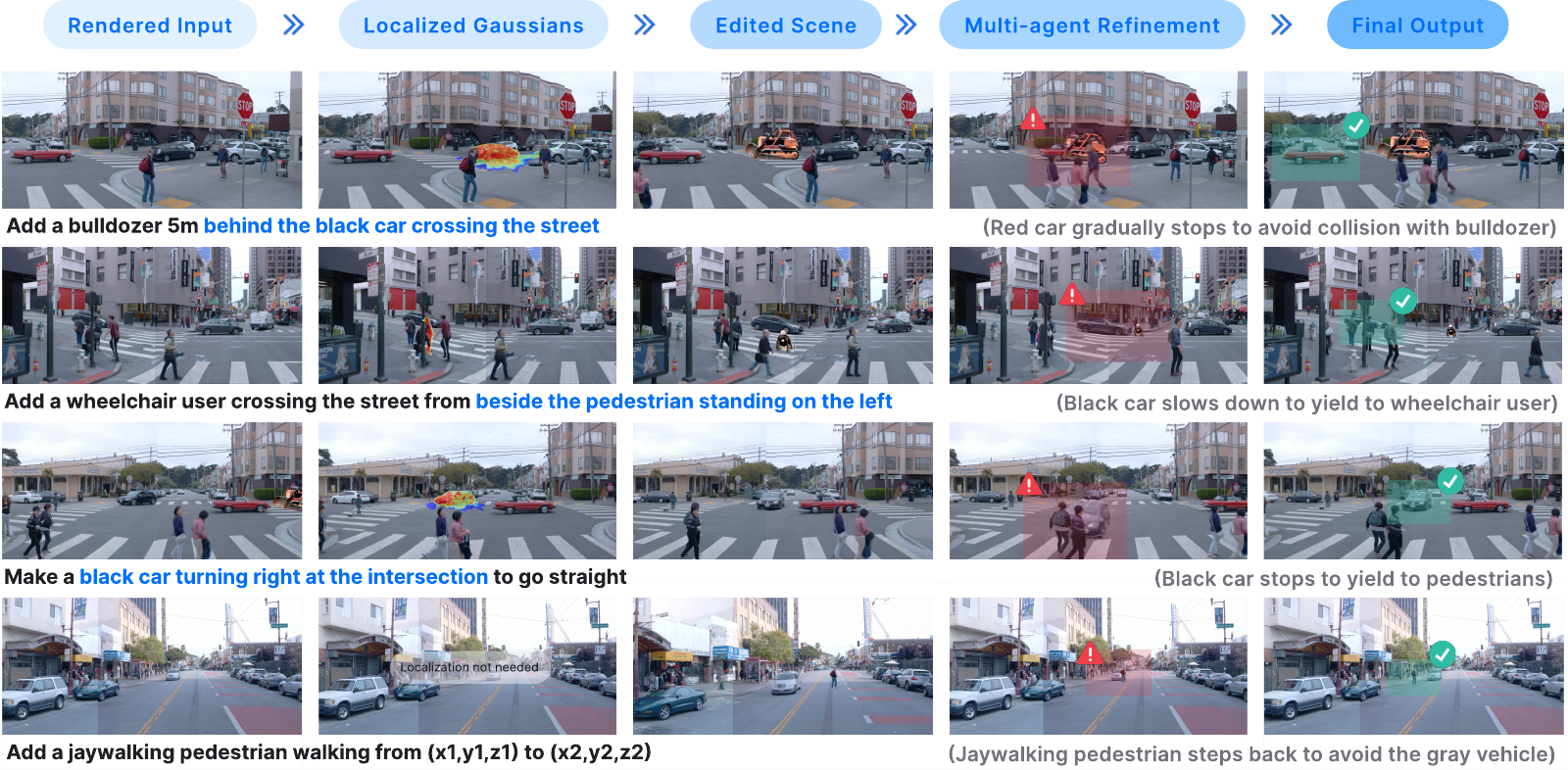}
\captionof{figure}{\textbf{Overview of SIMSplat.} Our framework enables language-guided, fine-grained editing in driving scenarios. Based on human/VLM-generated prompts, it first queries related objects from the Gaussian scene, then performs editing followed by multi-agent path refinement, ensuring the final scene remains realistic and consistent across all agents. SIMSplat supports a wide range of editing operations covering both vehicles and pedestrians. The \textcolor{red}{red}/\textcolor{ForestGreen}{green} areas indicate potential collision and refined regions.}
    \label{fig:language_alignment}
\end{figure*}

Driving simulators have long served as essential testbeds for autonomous driving. Traditional platforms such as CARLA~\cite{carla} and AirSim~\cite{shah2017airsimhighfidelityvisualphysical} offer scalability through game engines, while recent advances in neural scene reconstruction---including diffusion models~\cite{drivedreamer, drivedreamer4d, zhu2025scenecraftercontrollablemultiviewdriving, fan2024freesimfreeviewpointcamerasimulation}, Neural Radiance Fields~\cite{MARS, emernerf, guo2023streetsurfextendingmultiviewimplicit, chatsim}, and Gaussian Splatting~\cite{hugs, chen2025omnireomniurbanscene, zhou2024drivinggaussiancompositegaussiansplatting}---enable photo-realistic scene editing directly from sensor data. Language models have further been integrated to allow natural language control over scene manipulation. Yet despite this rapid progress, a fundamental gap persists: no existing framework provides a unified interface that seamlessly connects language understanding, object-level scene manipulation, and reactive multi-agent simulation.

This gap manifests as fragmentation across existing frameworks, where each addresses only a subset of the challenges. Language-guided editors~\cite{zhu2025scenecraftercontrollablemultiviewdriving, chatsim} depend on online 3D detectors at inference or given instance IDs/attributes to localize objects, translating natural language into rigid structured attributes. To edit objects, they rely on pre-stored virtual assets or simple removal, limiting the range of possible modifications on existing objects. Non-language editors~\cite{MARS, chen2025omnireomniurbanscene} support broader object-level editing but require predefined instance attributes for grounding and lack language controllability or automation. Across both lines of work, most frameworks focus exclusively on rigid objects such as vehicles while overlooking pedestrians despite their importance in safety-critical scenarios. Moreover, scenario feasibility is typically validated only for the ego vehicle or a single target object~\cite{drivedreamer4d, chatsim}, ignoring the cascading effects that any behavioral change has on surrounding agents. And to scale beyond individual hand-crafted scenarios, they require manual intervention at every step. No single framework addresses all of these challenges, as Table~\ref{tab:editing-comparison} summarizes.

We observe that these limitations share a common root: the absence of a \textit{shared scene representation} that is both expressive enough to capture diverse road dynamics and accessible enough to be controlled through natural language. If scene elements were represented in a structured, language-accessible form that jointly captures their visual appearance, dynamic behavior, and spatial context, then querying, editing, and simulating could all operate within a single coherent framework while enabling automation of the full pipeline. This observation motivates \textbf{SIMSplat}, which achieves this unification through scene-graph-based 4D Gaussian Splatting augmented with language-aligned features. Specifically, we embed three types of language features---appearance, motion, and location---directly into Gaussian scene-graph nodes, making the reconstructed scene queryable through free-form natural language. This \textit{language-Gaussian alignment} serves as the entry point to a fully integrated pipeline: language queries ground target objects within the Gaussian scene graph, edits operate on the same graph structure, and multi-agent refinement validates consistency across the scene.
Building on this language-accessible scene representation, SIMSplat enables a set of tightly integrated capabilities. Once target objects are grounded, the same scene-graph structure supports fine-grained manipulation through language commands, with a rich suite of object-level edits—including insertion, removal, replacement, and trajectory modification—covering both vehicles and pedestrians. A multi-agent path refinement module then refines trajectories of all scene-graph nodes, producing reactive simulations in which surrounding agents respond naturally to modifications at the full-scene level. Because language alignment bridges natural language to the underlying scene structure, the entire pipeline integrates directly with Vision-Language Models~(VLMs) for automated scenario mining. VLMs not only generate editing prompts but also evaluate rendered outputs for plausibility, iteratively refining asset selection and motion parameters until the modified scene satisfies both physical and semantic constraints---enabling scalable generation of long-tailed driving scenarios without manual intervention.

Experiments on the Waymo Open Dataset validate this approach. Our motion-aware language alignment more than doubles the object grounding accuracy of the strongest baseline. SIMSplat achieves the highest task completion rate across diverse editing categories and produces the lowest collision and failure rates in multi-agent scenarios. These results confirm that a language-aligned 4D Gaussian representation provides a sufficient and effective foundation for driving scene simulation, unifying capabilities that previously required manually coordinated systems.

In summary, we propose SIMSplat, a driving scene simulation framework built on 4D Gaussian scene graphs augmented with language-aligned features, enabling the following contributions:
\begin{itemize}
\item \textbf{Motion-aware language-Gaussian alignment}, which embeds appearance, motion, and location features into scene-graph Gaussians, enabling precise open-vocabulary grounding of dynamic road objects by jointly encoding trajectory dynamics through learned codebooks.
\item \textbf{Language-controlled object-level editor}, which builds on this language-aligned Gaussians to support fine-grained manipulation of both vehicles and pedestrians
---including safety-critical agents---through free-form natural language.
\item \textbf{Multi-agent path refinement}, which refines trajectories of all scene-graph nodes to propagate edits across all road participants, producing reactive simulations in which surrounding agents respond naturally to changes---validating feasibility at the full-scene level rather than for a single target.
\item \textbf{Automated scenario mining}, which bridges these capabilities through language, enabling VLMs to not only generate editing prompts but also evaluate rendered outputs and iteratively refine scene parameters, forming a closed-loop pipeline for scalable long-tailed scenario generation without manual intervention.
\end{itemize}

\begin{table*}[!t]
\caption{\textbf{Comparison of scene editors.} 
We compare prior simulators and scene editors across behavior editing, multi-agent path refinement, language control, automation, and object grounding, where \textit{Description} denotes free-form natural language requiring no predefined identifiers or spatial annotations.}
\centering
\setlength{\tabcolsep}{3pt}
\scriptsize
\begin{tabular*}{\textwidth}{@{\extracolsep{\fill}}l | c c | c c | c c c@{}}
\toprule
& \multicolumn{2}{c}{Editing} 
& \multicolumn{2}{c}{Path Refine.} 
& \multicolumn{3}{c}{Controllability} \\
\cmidrule(lr){2-3}\cmidrule(lr){4-5}\cmidrule(lr){6-8}
Method 
& \makecell{Ped.} 
& \makecell{Exist. Obj.} 
& \makecell{Vehicle} 
& \makecell{Ped.} 
& Language 
& Auto. 
& \makecell{Obj. Ground.} \\
\midrule
MARS \cite{MARS}           
& \redcross & \greencheck & \redcross & \redcross 
& \redcross & \redcross & Instance ID \\
HUGSIM \cite{hugsim}         
& \redcross & \greencheck & \greencheck & \redcross 
& \redcross & \greencheck & Tracking ID \\
ChatSim \cite{chatsim}       
& \redcross & \redcross & \redcross & \redcross 
& \greencheck & \redcross & Attributes \\
SceneCrafter \cite{zhu2025scenecraftercontrollablemultiviewdriving}   
& \redcross & \redcross & \redcross & \redcross 
& \greencheck & \greencheck & Bounding box \\
OmniRe \cite{chen2025omnireomniurbanscene}        
& \greencheck & \greencheck & \redcross & \redcross 
& \redcross & \redcross & Instance ID \\
\midrule
\textbf{Ours}   
& \greencheck & \greencheck & \greencheck & \greencheck 
& \greencheck & \greencheck & Description \\
\bottomrule
\end{tabular*}
\label{tab:editing-comparison}
\end{table*}
\section{Method}
\label{sec:method}

\begin{figure}[t]
    \centering
    \includegraphics[width=1.0\linewidth]{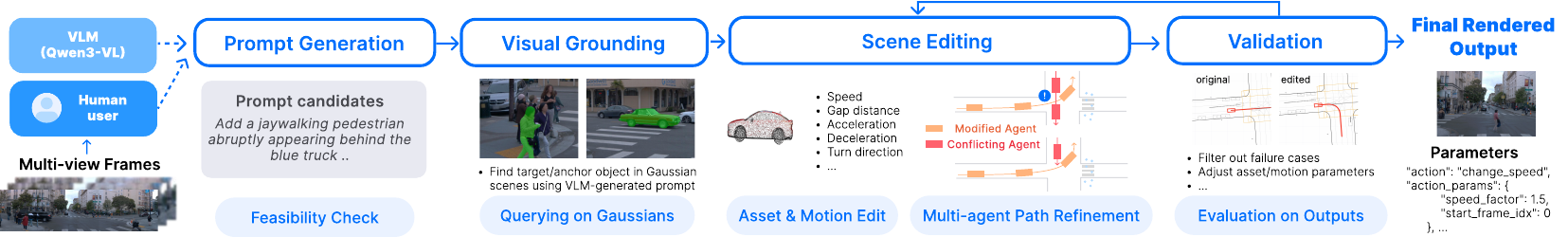}
\caption{\textbf{Editing process.} Given multi-view driving scenes, VLMs or humans generate prompts to perform the edit. After a feasibility check, target and anchor objects are localized in the language-grounded scene graph based on the prompt. Object editing is then applied, followed by multi-agent path refinement. The final output is rendered after adjusting parameters through iterative validation.}
    \label{fig:agent}
\end{figure}

\begin{figure*}[!t]
    \centering
    \includegraphics[width=1.0\textwidth]{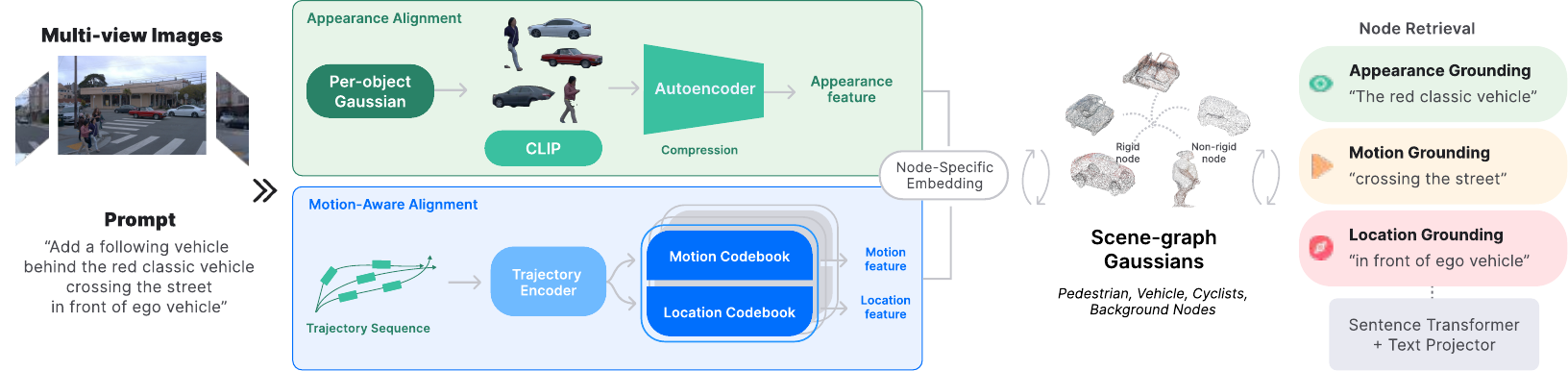}
\caption{\textbf{Pipeline of language alignment.} The appearance and motion-aware alignment modules extract appearance, motion, and location features, which are then embedded into scene-graph Gaussians. Given a natural language prompt, these features enable grounding of the corresponding objects in the road scene.}
    \label{fig:language_alignment}
\end{figure*}

SIMSplat is built on a scene-graph-based 4D Gaussian representation that is made language-accessible through learned feature alignment. We first describe the underlying scene-graph 4DGS formulation that provides structured, per-object Gaussian nodes (Section~\ref{sec:4dgs}), then introduce language-Gaussian alignment, which embeds appearance, motion, and location features into these nodes, making them queryable through free-form natural language (Section~\ref{sec:language_alignment}). Then, we describe how SIMSplat enables automated scenario mining (Section~\ref{sec:scenario_mining}) and how edited scenes are refined through multi-agent path prediction (Section~\ref{sec:path_refinement}). The editing process is illustrated in Figure~\ref{fig:agent}, and all mathematical notations are described in the supplementary material.

\subsection{Scene-Graph-based 4D Gaussian Splatting}
\label{sec:4dgs}
4DGS represents a scene as a set of anisotropic Gaussian blobs carrying spatial and appearance attributes. A Gaussian at time $t$ is represented as $g_i(t) = (\mu_i(t), s_i(t), q_i(t), c_i(t), o_i)$, where $\mu_i(t)$ is the 3D center, $s_i(t)$ the scale, $q_i(t)$ the orientation, $c_i(t)$ the color features, and $o_i$ the opacity of Gaussian $i$. When projected onto the image plane, each Gaussian contributes a 2D footprint determined by its position, scale, orientation, and opacity, and the final pixel color is obtained by alpha blending of overlapping Gaussians:
\begin{equation}
C_t(u) = \sum_{i} c_i(t),\alpha_i(t)\prod_{j<i}\big(1-\alpha_j(t)\big),
\label{eq:alpha_blend}
\end{equation}
where $\alpha_i(t)$ is the transparency weight of Gaussian $i$ at pixel $u$, computed from its spatial and opacity parameters. As shown in Eq.~\eqref{eq:alpha_blend}, the compositing follows a front-to-back blending rule. Extending beyond static 3DGS, the 4D formulation $\mathcal{G}(t) = \{ g_i(t) \}_{i=1}^N$ models temporal evolution by allowing Gaussians to transform smoothly over time.

To faithfully reconstruct dynamic scenes while preserving the motion and geometry of diverse road objects, we adopt a scene-graph formulation of 4DGS. In contrast to neural field-based approaches \cite{deformablegs}, the scene-graph decomposes a scene into separate nodes, enabling modular reconstruction and controllable editing. Following existing frameworks \cite{chen2025omnireomniurbanscene, zhou2024drivinggaussiancompositegaussiansplatting, hugs}, we represent the scene with three node types: rigid nodes (e.g., vehicles), non-rigid nodes (e.g., pedestrians), and the static background. Each node is defined in a canonical space and transformed into the world coordinate system at time $t$. For a rigid node $v$, the time-dependent Gaussians are given by $G^{\text{rigid}}_v(t) = T_v(t) \otimes \bar{G}^{\text{rigid}}_v$, where $\bar{G}^{\text{rigid}}_v$ are canonical Gaussians and $T_v(t) \in SE(3)$ is the rigid body transformation. For a non-rigid node $h$, the Gaussians undergo both global motion and local deformation, expressed as $G^{\text{nonrigid}}_h(t) = T_h(t) \otimes F\big(\bar{G}^{\text{nonrigid}}_h, t\big)$, where $F(\cdot, t)$ denotes a non-rigid deformation function applied to the canonical Gaussians, modeling human pose and gestures through joint-level prediction.

Finally, the complete dynamic scene at time $t$ is obtained by aggregating all node types into a unified scene graph $\mathcal{S}(t) = \{ G_{\text{bg}}, \{ G^{\text{rigid}}_v(t) \}, \{ G^{\text{nonrigid}}_h(t) \} \}$, where $G_{\text{bg}}$ represents static background Gaussians, and the other terms correspond to rigid and non-rigid nodes. The final rendered scene is produced through blending and rasterization of all Gaussian components.

\subsection{Language-Gaussian Alignment}
\label{sec:language_alignment}
To make the scene-graph representation controllable through natural language, we introduce language-Gaussian alignment, which embeds language semantics directly into 4D object nodes. As prompts often involve spatial and temporal understanding of road dynamics, accurately localizing target objects according to the prompt within dense driving scenes is challenging. For example, simple attribute matching (e.g., color, type) is ill-suited for complex driving environments where such attributes frequently overlap across objects. Bounding box prompting, on the other hand, is prone to localization errors, especially when generated directly from VLMs, which tend to produce imprecise spatial coordinates. To address these issues, rather than relying on prior information or ambiguous matching, we embed appearance-, motion-, and location-language features directly into the Gaussians, enabling flexible localization solely from free-form descriptive queries. This language grounding supports more accurate and generalizable object retrieval at both the pixel and node level.

\noindent \textbf{Appearance Alignment.} As shown in Figure \ref{fig:language_alignment}, first, we embed static language features on Gaussians to capture object appearance in a time-independent manner. We extract segmentation masks $M_t(o)$ for each object $o$ at time $t$ by projecting per-node Gaussians from the scene graph onto the image plane, removing the need for an external segmentation model at this stage. From each masked object, we extract CLIP features $f^{\text{CLIP}}_t(o) \in \mathbb{R}^D$. Since CLIP features are high-dimensional, we train a lightweight autoencoder $(E,\Psi)$ to compress them into a latent space:  
\begin{equation}
h_t(o) = E(f^{\text{CLIP}}_t(o)) \in \mathbb{R}^d, \quad d \ll D,
\label{eq:clip_autoencoder}
\end{equation}
with the reconstruction objective $\Psi(h_t(o)) \approx f^{\text{CLIP}}_t(o)$. As defined in Eq.~\eqref{eq:clip_autoencoder}, the encoder $E(\cdot)$ produces low-dimensional latent codes $h_t(o)$, which act as ground-truth supervision for Gaussian appearance features $c_i^{\text{app}} \in \mathbb{R}^d$, optimized in a self-supervised manner. Each Gaussian at time $t$ is therefore extended as
\begin{equation}
g_i(t) = \big(\theta_i(t), c_i^{\text{app}}\big),
\label{eq:appearance_gaussian}
\end{equation}
where $\theta_i(t) = (\mu_i(t), s_i(t), q_i(t), c_i(t), o_i)$ denotes the original Gaussian parameters, and the additional term $c_i^{\text{app}}$ represents the appearance-aligned feature in the latent space. As shown in Eq.~\eqref{eq:appearance_gaussian}, this formulation integrates language-aligned appearance information directly into the Gaussian representation. During inference, the rasterized features $c_i^{\text{app}}$ are decoded back into the CLIP space $\Psi(c_i^{\text{app}}) \in \mathbb{R}^D$ and compared with encoded text queries, enabling open-vocabulary selection at both pixel and Gaussian levels.

\noindent \textbf{Motion-Aware Alignment.} Through experiments, we found that existing methods to embed static or dynamic language features, such as LangSplat and 4DLangSplat~\cite{qin2024langsplat3dlanguagegaussian,li20254dlangsplat4dlanguage}, are not capable of understanding the behavior of road objects. In particular, queries often involve driving-scene-specific descriptions such as ``a pedestrian crossing the street on the left side of the ego vehicle'' or ``a vehicle crossing the intersection from right to left,'' which include motion cues (e.g., speed, direction, turning) and relative location. However, prior methods primarily capture object appearance (e.g., color, shape) and are limited to simple indoor dynamics (e.g., opening or closing a cup).

To address this limitation, we propose a motion-aware alignment module that encodes trajectories $X = \{(x_t, y_t)\}_{t=1}^T$ of each object into a latent representation $z = E_{\text{traj}}(X) \in \mathbb{R}^d$ using a trajectory encoder, which is then associated with two codebooks: a \textit{motion codebook} and a \textit{location codebook}, defined as
\begin{equation}
\mathcal{C}^{\text{mot}} = \{p_k^{\text{mot}}\}_{k=1}^{K_m},\quad
\mathcal{C}^{\text{loc}} = \{p_k^{\text{loc}}\}_{k=1}^{K_l},
\label{eq:codebooks}
\end{equation}
where each prototype vector $p$ corresponds to a typical motion or location on the road (e.g., turning left, moving right to left, in front of ego, on the left side of ego). To account for the different motion patterns of rigid and non-rigid agents, we maintain separate motion codebooks for vehicles and pedestrians, while sharing the same location codebook across object types. Given a trajectory embedding $z$, similarity scores with prototypes are computed, and the aggregated features $f^{\text{mot}}, f^{\text{loc}} \in \mathbb{R}^d$ are obtained as convex combinations of the corresponding codebook vectors. These serve as temporal language features capturing both the object's movement and location relative to the ego view.

During training, motion and location alignment losses are combined with commitment terms as $\mathcal{L}_{\text{temp}} = \lambda_{\text{align}}(\mathcal{L}_{\text{mot}} + \mathcal{L}_{\text{loc}}) + \lambda_{\text{commit}}\,\mathcal{L}_{\text{commit}}$, where $\mathcal{L}_{\text{mot}}$ and $\mathcal{L}_{\text{loc}}$ measure the cosine distance with text prototypes, and the commitment terms $\mathcal{L}_{\text{commit}}$ enforce closeness between $z$ and the weighted codebook features.

Finally, the temporal feature is associated with each object node, represented as  
\begin{equation}
G_o(t) = \big(\{g_i(t)\}_{i=1}^{N_o},\, c_o^{\text{temp}}\big),
\label{eq:temporal_node}
\end{equation}
where $\{g_i(t)\}_{i=1}^{N_o}$ are the Gaussians belonging to object $o$, and $c_o^{\text{temp}} = (f^{\text{mot}}, f^{\text{loc}})$
 encodes the behavioral context at the object level. As shown in Eq.~\eqref{eq:temporal_node}, this formulation binds temporal features to each object node, enabling open-vocabulary reasoning over object dynamics. During inference, object trajectories are first encoded into temporal features $c_o^{\text{temp}}$ and compared with the encoded text prompt. This yields a unified embedding that compactly represents the behavioral dynamics of each object. The detailed implementation of motion-aware alignment can be found in the supplementary material.

\noindent \textbf{Gaussian Object Querying.} For querying the scene with both appearance and temporal features, we first encode the text prompt with E5 Sentence Transformer \cite{e5}. Appearance features rendered from Gaussians are decoded through a pretrained autoencoder and compared against prompt embeddings to generate pixel-level similarity maps, which are post-processed into binary masks. These masks are then used to select candidate object instances per prompt. For vehicles and pedestrians, separate trajectory encoders are loaded (trained with motion and location codebooks) together with their text projectors, and the extracted temporal features of candidate trajectories are compared with prompt embeddings using cosine similarity. At inference, appearance similarity first narrows down candidate regions, after which motion and location features are combined to select the most relevant object based on the highest similarity.

\subsection{Automated Scenario Mining}
\label{sec:scenario_mining}
While human-guided driving scene editing enables fine-grained control, it inherently limits scalability and diversity due to its reliance on manual prompting and intervention. To address this, SIMSplat supports VLM-driven scenario mining that automates scene augmentation at scale, integrating scenario generation, validation, and refinement within a single pipeline.

As shown in Figure~\ref{fig:agent}, the VLM takes multi-view driving scene images as input and generates natural language prompts describing potential scene modifications. Each prompt is validated against a predefined set of supported editing functions to ensure feasibility. SIMSplat then performs visual grounding in the language-aligned 4D Gaussian scene to identify the referenced objects. For example, given the prompt ``Make the black car turning at the intersection go straight,'' the VLM extracts the caption ``black car turning at the intersection'' and retrieves the most relevant object from the Gaussian scene, as described in the previous section. Because language semantics are embedded directly into the Gaussian scene-graph nodes, object localization becomes an intrinsic property of the representation, eliminating the need for online detectors or manually specified spatial annotations.

SIMSplat then designs object trajectories by adjusting motions (e.g., accelerating, reversing, lane changing, or turning), inserting new dynamic or static objects (e.g., adding a following vehicle or placing a roadside obstacle), or performing object replacement and removal. The framework supports flexible action parameterization (including speed, direction, start time, relative distance, and start/end positions) to enable diverse and fine-grained scene edits. Generated motions are further validated and refined by the multi-agent path refinement module to ensure global consistency, as discussed in the following section.

For new object insertion, SIMSplat retrieves appropriate assets from an asset bank covering vehicles, pedestrians, traffic signs, and road objects such as barriers, cones, and construction equipment. Unlike conventional simulators that rely on synthetic or static pedestrian models, our asset bank includes dynamic real-world pedestrian assets extracted directly from sensor data, each annotated with descriptive captions (e.g., ``a pedestrian walking from left to right with a red backpack''), enabling realistic insertion and improving simulation fidelity.

After editing, the modified 3D scene is rendered onto the 2D image plane and validated. Invalid outcomes such as collisions or off-road behavior are first filtered out. The VLM then evaluates semantic plausibility by taking the edited RGB images, BEV map, and trajectories of target and surrounding agents as input to determine whether the modification faithfully follows the original prompt. If inconsistencies are detected or plausibility is not satisfied, the system iteratively refines asset selection or motion parameters. Once all constraints are met, the editing process is finalized. While the framework supports full automation, human intervention is also available for manual prompting and detailed parameter tuning.

\subsection{Multi-agent Path Refinement} 
\label{sec:path_refinement}
While the trajectory outputs generated by pre-defined function sets provide a useful starting point, they are not specifically designed as a multi-agent path planner and therefore have limited ability to reason about surrounding traffic participants and road context. Existing approaches often generate modified trajectories with built-in functions, but these typically focus only on producing collision-free paths for the target object without considering the impact of the modification on other agents. To address this limitation, our framework aims to create reactive scenarios in which not only the edited target agents but also the surrounding agents respond naturally to changes.

To achieve this, we apply multi-agent path refinement using a motion prediction model that forecasts the future trajectories of all agents potentially affected by the edits. Specifically, we train SMART-1B \cite{wu2024smartscalablemultiagentrealtime} on the Waymo Open Dataset \cite{waymo} and use it to simulate the collective behavior of road agents. Given the pre-generated trajectory $\tau^{\text{edit}}$ of the target object from the Motion Controller and the observed history $X_{1:t} = \{x^o_{1:t}\}_{o=1}^N$ of all agents, the predictor $P$ generates future rollouts
\begin{equation}
\hat{X}_{t+1:T} = P(X_{1:t}, \tau^{\text{edit}}),
\label{eq:motion_refine}
\end{equation}
where $\hat{X}_{t+1:T}$ denotes the refined trajectories for both the target and surrounding objects. This refinement enables our framework to roll out realistic multi-agent scenarios. For instance, a following vehicle may detour or stop when the edited vehicle ahead brakes abruptly, or adjacent cars may react to avoid a collision when a jaywalking pedestrian appears. This process not only adapts vehicle behaviors but also captures pedestrian dynamics, e.g., a pedestrian halts when a newly inserted obstacle blocks the crosswalk. Since the refinement depends on a prediction model and may be sensitive to uncertainty or failure cases, we also provide the option to bypass this module when deterministic editing is preferred.

\section{Experiments}

\subsection{Datasets and Settings}
To evaluate the proposed driving scene editor in realistic driving environments, we conduct experiments on the Waymo Open Dataset \cite{waymo}. For training, we use images from the front, front-left, and front-right cameras, with 100 frames sampled at 10 Hz for each sequence. Every 10th frame is reserved for the test set, and training is performed on a single NVIDIA A100 GPU. For motion generation, the first 11 timesteps of each trajectory are used as input, which predicts 80 future timesteps. For VLM-driven automation, Qwen3-VL-32B \cite{bai2025qwen3vltechnicalreport} is used. Further training and evaluation details are provided in the supplementary material.

\subsection{Road Object Querying}
\label{main:querying}

\begin{figure}[t]
\centering
\begin{minipage}[t]{0.48\linewidth}
    \vspace{0pt} 
    \centering
    \includegraphics[width=\linewidth]{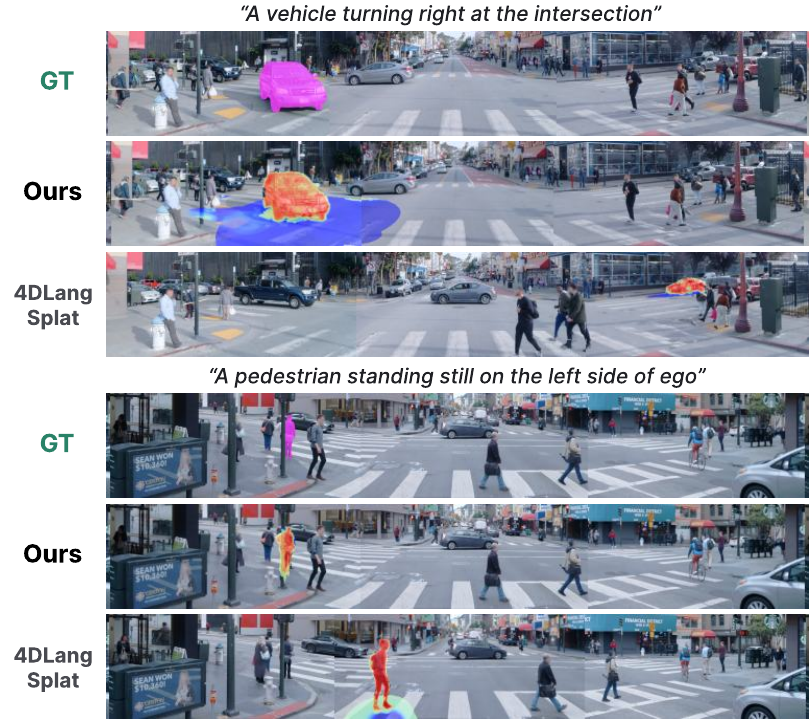}
    \captionof{figure}{\textbf{Qualitative comparison of object querying.} Our method effectively captures the motion and location of road agents within the scene.}
    \label{fig:querying}
\end{minipage}
\hfill
\begin{minipage}[t]{0.48\linewidth}
    \centering
    \small
\captionof{table}{\textbf{Object querying results.} Metrics reported are accuracy (Acc.) and video-level IoU (vIoU).}
\resizebox{0.95\linewidth}{!}{%
\begin{tabular}{l|c c c c c c}
  \toprule
  Method & \multicolumn{2}{c}{Vehicle} & \multicolumn{2}{c}{Pedestrian} & \multicolumn{2}{c}{Total} \\
  \cmidrule(lr){2-3} \cmidrule(lr){4-5} \cmidrule(lr){6-7}
   & Acc. & vIoU & Acc. & vIoU & Acc. & vIoU\\
  \midrule
  LangSplat & 0.27 & 0.47 & 0.30 & 0.32 & 0.28 & 0.40 \\
  4DLangSplat & 0.33 & 0.53 & 0.37 & 0.44 & 0.35 & 0.49 \\
  \specialrule{0.5pt}{1pt}{1pt}
  Ours & \textbf{0.83} & \textbf{0.81} & \textbf{0.63} & \textbf{0.78} & \textbf{0.73} & \textbf{0.80} \\
  \bottomrule
\end{tabular}
}
\label{tab:querying-results}
    \captionof{table}{\textbf{Ablations on language feature components.} \textit{App.}, \textit{Mot.}, and \textit{Loc.} denote \textit{Appearance}, \textit{Motion}, and \textit{Location}.}
\resizebox{\linewidth}{!}{%
\begin{tabular}{c c c|c c c c c c}
  \toprule
  \multicolumn{3}{c}{Features} &
  \multicolumn{2}{c}{Vehicle} &
  \multicolumn{2}{c}{Pedestrian} &
  \multicolumn{2}{c}{Total} \\
  \cmidrule(lr){1-3}
  \cmidrule(lr){4-5}
  \cmidrule(lr){6-7}
  \cmidrule(lr){8-9}
  App. & Mot. & Loc. &
  Acc. & vIoU &
  Acc. & vIoU &
  Acc. & vIoU \\
  \midrule
  \checkmark & & &
  0.53 & 0.56 &
  0.27 & 0.38 &
  0.40 & 0.47 \\
  \checkmark & \checkmark & &
  0.73 & 0.70 &
  0.27 & 0.38 &
  0.50 & 0.54 \\
  \checkmark & & \checkmark &
  0.57 & 0.64 &
  0.33 & 0.42 &
  0.45 & 0.53 \\
  \specialrule{0.5pt}{1pt}{1pt}
  \checkmark & \checkmark & \checkmark &
  \textbf{0.83} & \textbf{0.81} &
  \textbf{0.63} & \textbf{0.78} &
  \textbf{0.73} & \textbf{0.80} \\
  \bottomrule
\end{tabular}
}
    \label{tab:querying-ablation}
\end{minipage}
\end{figure}

We first verify that language-aligned Gaussians provide accurate object grounding by comparing open-vocabulary querying results against state-of-the-art baselines, LangSplat and 4DLangSplat. As shown in Table \ref{tab:querying-results}, when given queries containing object descriptions, our method achieves an overall accuracy of 0.73 and a vIoU of 0.80, outperforming other models by a significant margin. In particular, for vehicle objects, our method attains 0.83 in accuracy and 0.81 in vIoU, nearly doubling the performance of 4DLangSplat. Even for pedestrians, which are typically smaller and harder to detect than vehicles, our model achieves 70.3\% and 77.3\% higher accuracy and vIoU scores, respectively, compared to 4DLangSplat.

Figure \ref{fig:querying} illustrates qualitative examples of querying results. Compared with 4DLangSplat, which often fails to locate the correct object following a motion description, our model successfully identifies the target. For instance, in the first case, our model successfully finds the vehicle turning right among several moving vehicles, whereas 4DLangSplat instead selects a vehicle driving straight. In another case, our model identifies the pedestrian standing still on the left side, while 4DLangSplat incorrectly highlights a pedestrian walking across the crosswalk. Although not shown in the figure, even when specific appearance-based prompts are provided (e.g., black sedan, pedestrian with red coat), multiple agents with similar appearances may exist due to the density of road participants. In such cases, our model resolves ambiguity by incorporating motion and location cues into the language field. These results demonstrate that our motion-aware alignment strategy effectively embeds the dynamic nature of road agents, thereby enabling more accurate and reliable language-grounded querying. 

To examine the contribution of motion and location features in our alignment module, we conduct ablation studies on different language feature components. As shown in Table~\ref{tab:querying-ablation}, combining motion and location features with appearance features yields the best performance across all metrics and tasks. Moreover, when motion features are used alongside appearance features, the accuracy and vIoU for vehicle objects improve, whereas combining location features with appearance features enhances the metrics for pedestrian objects. These results indicate that when applied separately, motion and location features each contribute distinct benefits to different object types, and their combination significantly enhances the overall querying performance for dynamic road objects by providing richer behavioral cues.

\subsection{Scene Editing}

\label{main:scene_editing}
We evaluate scene editing results across various types of modifications and language commands. As baselines, we primarily compare against ChatSim \cite{chatsim}, which provides the most diverse range of language-based editing among existing frameworks, and OmniRe \cite{chen2025omnireomniurbanscene}, which supports object-level editing in Gaussian scenes. For fairness, all models are trained under the same scenario. 

\begin{figure*}[t]
    \centering
    \includegraphics[width=\linewidth]{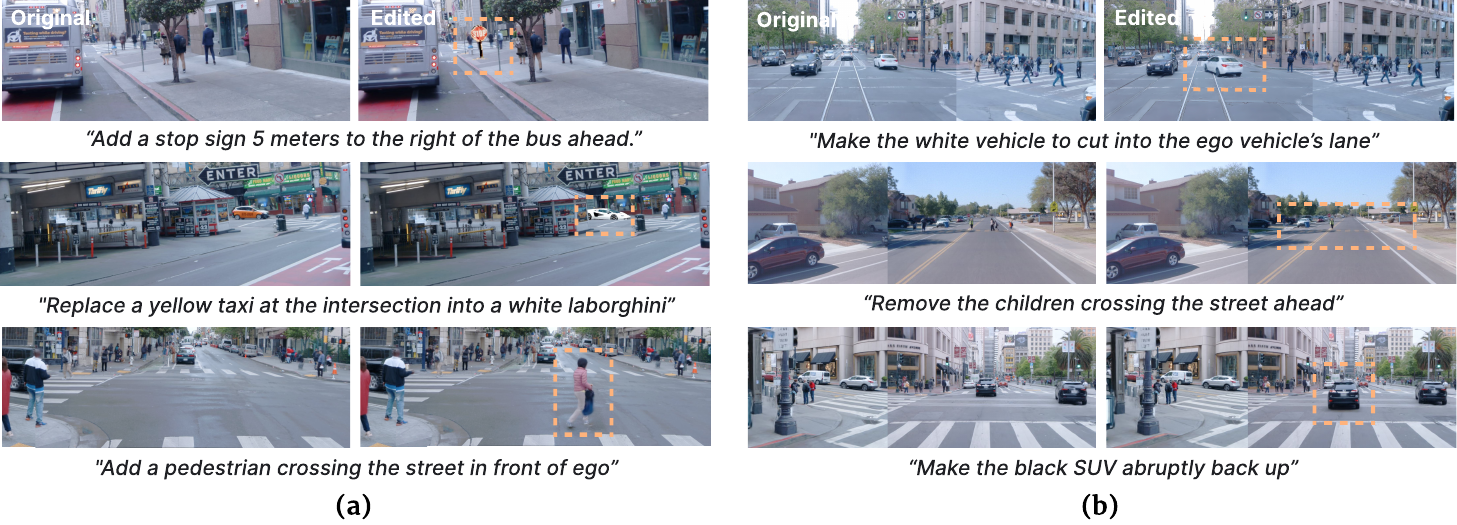}
\caption{\textbf{Qualitative editing results.} (a) Human-guided editing. (b) VLM-driven editing. SIMSplat supports detailed modifications including vehicles, pedestrians, and static road objects. Summarized versions of the prompts are shown for better visibility.}
    \label{fig:tasks}
\end{figure*}

\noindent \textbf{Qualitative Results.} Figure~\ref{fig:tasks} presents qualitative examples of scene editing using our framework, with human-guided editing shown on the left and VLM-driven editing on the right. As illustrated in the prompts, object placement can be specified either through relative descriptions (e.g., ``in front of the construction worker'' or ``next to the bus ahead'') or via explicit coordinates $(x, y, z)$. Beyond editing individual objects, our framework also supports group-level commands, such as ``remove all crossing pedestrians/vehicles'' Notably, as shown in the bottom-left example, our framework uniquely supports pedestrian editing. It enables not only modification of existing pedestrians (e.g., adjusting walking speed) but also insertion of real-human pedestrian assets extracted from other scenarios. This capability allows users to introduce realistic pedestrians while preserving natural joint motions and gestures. For instance, the figure demonstrates a newly inserted pedestrian crossing the street with coherent and natural movement. Such functionality further facilitates the creation of safety-critical scenarios, e.g., ``insert a jaywalking pedestrian'' or ``a wheelchair user abruptly appearing behind the truck,'' through precise and controllable pedestrian manipulation. In addition to pedestrian editing, our framework provides detailed vehicle control, including speed adjustment and trajectory modification (e.g., turning left, changing lanes, or reversing). As shown in (b), these behavior changes are effectively used to generate adversarial scenarios such as aggressive cut-ins or abrupt reversals. Motion edits can be further specified using either absolute or relative positioning (e.g., ``start 2m next to the white car,'' ``stop at $(x, y, z)$''), and new vehicles can likewise be introduced with user-defined behaviors.

\begin{figure*}[!t]
    \centering
    \includegraphics[width=1.0\textwidth]{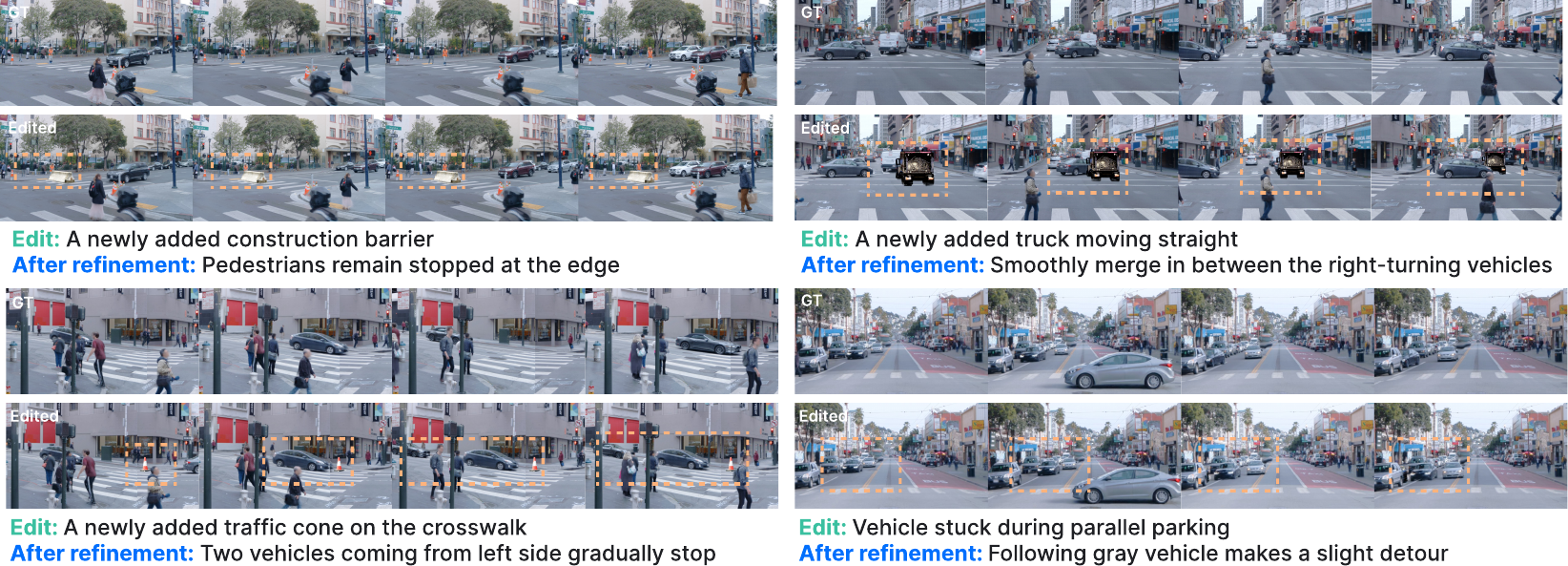}
    \caption{\textbf{Qualitative results with predictive path refinement.} The multi-agent path refinement adapts both target and surrounding objects to interact in edited scenes.}
    \label{fig:refine_results}
\end{figure*}

Figure~\ref{fig:refine_results} showcases how our framework refines agent trajectories and integrates predictive interactions into edited scenes. In the top-left case, a construction worker and a pedestrian standing on the left originally cross the street, but after a barrier blocks the crosswalk, they remain stopped at the edge. In the top-right case, we insert a truck proceeding straight through an intersection. To merge with right-turning vehicles, it waits for a safe gap before merging smoothly into traffic, demonstrating our framework's ability to adapt to new objects in ongoing traffic conditions. In the bottom-left case, a traffic cone is added in the middle of a crosswalk. Two approaching vehicles detect the new obstacle and gradually stop in front of it. In the last case, a vehicle attempting parallel parking abruptly halts, likely due to becoming stuck. The following gray vehicle then makes a slight detour to avoid it. As shown in these examples, multi-agent path refinement adapts all agents, including both vehicles and pedestrians, to coordinately respond to changes. These results demonstrate that our framework not only enables efficient scene editing but also generates coherent multi-agent interactions through predictive path refinement.

\noindent \textbf{Task Completion.} We evaluate task completion rates using prompts grouped into three categories: adding new objects, modifying existing objects, and removal. As shown in Table~\ref{tab:task-results}, our human-guided model achieves the highest overall completion rate (87.2\%). Even without human intervention, the VLM-driven version outperforms all baseline methods, demonstrating the effectiveness of our automated pipeline built upon language-aligned Gaussians. ChatSim supports the insertion of new vehicles and static objects, but does not enable pedestrian editing or modification of existing objects, as it does not model individual road agents explicitly nor capture their temporal motion dynamics. In contrast, OmniRe allows adding pedestrians and modifying existing objects through its scene-graph-based reconstruction. However, it lacks the ability to control objects with given language commands, such as specifying target objects or relative positions. Moreover, ChatSim identifies existing objects primarily through attribute matching, whereas our model can localize objects using free-form language descriptions. This capability aligns particularly well with the VLM-leveraged framework, which excels at generating rich semantic descriptions rather than explicit bounding box or pixel-level selection. Overall, SIMSplat achieves substantially higher task completion rates across diverse editing categories, owing to its accurate object grounding within the 4D scene and extensive editing support for both vehicles and pedestrians.

\begin{table}[t]
    \caption{\textbf{Task completion results.} The metric represents the success rate of executed prompts. Obj. denotes a static object. $^\dagger$ denotes an extended version with editing functions.}
    \centering
    \renewcommand{\arraystretch}{1.0}
    \setlength{\tabcolsep}{6pt}
    \resizebox{0.8\linewidth}{!}{%
    \begin{tabular}{l|c c c c c c c}
      \toprule
      \multirow{2}{*}{Method} &
      \multicolumn{3}{c}{Add new} &
      \multicolumn{2}{c}{Modify} &
      \multirow{2}{*}{Remove} &
      \multirow{2}{*}{Total} \\
      \cmidrule(lr){2-4} \cmidrule(lr){5-6}
       & Vehicle & Stat. Obj. & Ped. & Vehicle & Ped. & & \\
      \midrule
        ChatSim \cite{chatsim} & 46.7 & 63.3 & 0.0 & 0.0 & 0.0 & 50.0 & 26.7 \\
        OmniRe$^\dagger$\cite{chen2025omnireomniurbanscene} & 26.7 & 23.3 & 16.7 & 36.7 & 23.3 & 40.0 & 27.8 \\
        \midrule
        Ours (VLM) & 70.0 & 70.0 & 66.7 & 73.3 & 66.7 & 80.0 & 71.1 \\
        Ours (human) & \textbf{86.7} & \textbf{83.3} & \textbf{93.3} & \textbf{83.3} & \textbf{90.0} & \textbf{86.7} & \textbf{87.2} \\
      \bottomrule
    \end{tabular}}
    \label{tab:task-results}
\end{table}

\begin{table}[t]
\centering
\small
\begin{minipage}[t]{0.49\linewidth}
    \captionof{table}{\textbf{Motion generation results.} We report collision rates, off-road driving, and total failures.}
    \centering
    \renewcommand{\arraystretch}{1.0}
    \resizebox{\linewidth}{!}{%
    \begin{tabular}{l|c c c c}
      \toprule
      Method & Col. veh. & Col. ped. & Off-road & Failure \\
      \midrule
    ChatSim & 53.3 & 36.7 & 33.3 & 53.3 \\
    OmniRe$^\dagger$ & 63.3 & 30.0 & 26.7 & 70.0 \\
    GPT2Motion & 60.0 & 73.3 & 36.7 & 80.0 \\
    \midrule
    Ours (w/o ref.) & 43.3 & 26.7 & 10.0 & 53.3 \\
    Ours & \textbf{13.3} & \textbf{16.7} & \textbf{6.7} & \textbf{20.0} \\
      \bottomrule
    \end{tabular}}
    \label{tab:motion-results}
\end{minipage}
\hfill
\begin{minipage}[t]{0.49\linewidth}
    \captionof{table}{\textbf{Comparison of VLMs.} We report the number of iterations, task completion, and plausibility score.}
    \centering
    \renewcommand{\arraystretch}{1.0}
\resizebox{0.95\linewidth}{!}{%
\begin{tabular}{l|ccc}
  \toprule
  Model & \# Iter. & Task Comp. & Plaus. \\
  \midrule
  Qwen3-VL-32B \cite{bai2025qwen3vltechnicalreport} & 4.5 & 71.1 & 2.80 \\
  Qwen3-VL-8B \cite{bai2025qwen3vltechnicalreport}  & 7.4 & 48.3 & 2.10 \\
  GPT-5         & \textbf{3.6} & 78.9 & \textbf{3.25} \\
  GPT-5-mini    & 4.1 & 73.3 & 2.95 \\
  Claude-Opus-4.1 & 3.9 & \textbf{80.0} & 3.10 \\
  \bottomrule
\end{tabular}}
    \label{tab:vlm-results}
\end{minipage}
\end{table}

\noindent \textbf{Motion Generation.} We then assess whether multi-agent path refinement produces consistent simulations after editing. Table~\ref{tab:motion-results} reports failure rates in terms of interactions with the surrounding environment. GPT2Motion refers to using GPT-5 directly as a motion generator, and we also evaluate a variant of our model without the multi-agent path refinement module. As shown, when analyzing total failure cases, our model achieves the lowest error rates. Other baselines perform well when a static object is placed in isolation, away from surrounding agents. However, when the edited object must coordinate with nearby agents, collisions or off-road driving occur much more frequently. While target vehicle-oriented planning and validation may succeed in sparse traffic or isolated settings, they often fail to adapt effectively in complex urban scenarios. These results highlight that scene editing in real-world driving requires not only generating collision-free trajectories for a single object but also dynamically adjusting the motions of surrounding agents in line with their predicted interactions.

\noindent \textbf{VLM-driven Scenario Mining.} Finally, we compare how different VLMs perform when driving the automated pipeline by evaluating the number of iterations, task completion rate, and plausibility score across open-source and commercial VLMs. The number of iterations reflects how efficiently a model converges to its final output. As shown in Table~\ref{tab:vlm-results}, GPT-5 generates scenes with the fewest iterations (3.6 on average), whereas Qwen3-VL-8B requires the most. Larger models such as GPT-5 and Claude-Opus-4.1 also achieve higher task completion rates of 78.9\% and 80.0\%, respectively. The plausibility score is evaluated using Gemini, following prior work in 3D generative modeling that leverages VLM-based evaluators to quantify output quality~\cite{huang2025fireplacegeometricrefinementsllm, wu2024gpt4visionhumanalignedevaluatortextto3d}. Scores range from 1 (poor) to 4 (good), with detailed evaluation rubrics provided in the supplementary material. Consistent with the other metrics, GPT-5 and Claude-Opus-4.1 achieve the highest plausibility scores of 3.25 and 3.10, respectively. Although SIMSplat currently employs the open-source Qwen3-VL-32B model, our results suggest that larger VLMs can further improve both the efficiency and quality of scenario mining, while smaller models remain attractive for their faster inference speed and lower computational cost. 

\section{Conclusion}
We presented SIMSplat, a framework that bridges language understanding, object-level scene editing, and reactive multi-agent simulation within a single pipeline built on language-aligned 4D Gaussian scene graphs. By embedding appearance, motion, and location features into scene-graph Gaussians, SIMSplat makes reconstructed driving scenes queryable through free-form natural language, bridging language understanding to object-level editing and multi-agent simulation. Our motion-aware alignment more than doubles the grounding accuracy of prior methods, and this language-aligned scene-graph in turn enables a rich suite of edits---including fine-grained pedestrian manipulation---while multi-agent path refinement ensures that surrounding agents respond naturally to changes. The full pipeline integrates with VLMs for automated scenario mining, achieving the highest task completion rate and the lowest failure rate across diverse driving scenarios, with larger VLMs further improving both efficiency and output quality.


%
%
\bibliographystyle{splncs04}
\bibliography{main}
\end{document}